\documentclass[sigconf]{acmart}

\AtBeginDocument{%
  \providecommand\BibTeX{{%
    \normalfont B\kern-0.5em{\scshape i\kern-0.25em b}\kern-0.8em\TeX}}}
\copyrightyear{2020} 
\acmYear{2020} 
\setcopyright{acmcopyright}
\acmConference[IWCTS'20]{13th International Workshop on Computational Transportation Science }{November 3, 2020}{Seattle, WA, USA}
\acmBooktitle{13th International Workshop on Computational Transportation Science (IWCTS'20), November 3, 2020, Seattle, WA, USA}
\acmPrice{15.00}
\acmDOI{10.1145/3423457.3429361}
\acmISBN{978-1-4503-8166-6/20/11}

\usepackage{amsmath}
\usepackage{graphicx}
\usepackage{textcomp}
\usepackage{xcolor}
\usepackage{hyperref}       
\usepackage{url}            
\usepackage{booktabs}       
\usepackage{amsfonts}
\usepackage{nicefrac}       
\usepackage{microtype}      
\usepackage{caption}
\usepackage{subcaption}
\usepackage{algorithm}
\usepackage[noend]{algpseudocode}
\usepackage{gensymb}

\begin{document}

\title[VAE-Info-cGAN]{VAE-Info-cGAN: Generating Synthetic Images by Combining Pixel-level and Feature-level Geospatial Conditional Inputs}

\author{Xuerong Xiao}
\authornote{Equally contributed to the paper.}
\email{xuerong@stanford.edu}
\affiliation{%
    \institution{Stanford University}
    \city{Stanford}
    \state{CA}
    \postcode{94305}
}
\author{Swetava Ganguli}
\authornotemark[1]
\authornote{Corresponding Author. Alternative EMail: swetava@cs.stanford.edu}
\email{swetava@apple.com}
\affiliation{%
    \institution{Apple}
    \city{Cupertino}
    \state{CA}
    \postcode{95014}
}
\author{Vipul Pandey}
\email{vipul@apple.com}
\affiliation{%
    \institution{Apple}
    \city{Cupertino}
    \state{CA}
    \postcode{95014}
}

\begin{abstract}
Training robust supervised deep learning models for many geospatial applications of computer vision is difficult due to dearth of class-balanced and diverse training data. Conversely, obtaining enough training data for many applications is financially prohibitive or may be infeasible, especially when the application involves modeling rare or extreme events. Synthetically generating data (and labels) using a generative model that can sample from a target distribution and exploit the multi-scale nature of images can be an inexpensive solution to address scarcity of labeled data. Towards this goal, we present a deep conditional generative model, called VAE-Info-cGAN, that combines a Variational Autoencoder (VAE) with a conditional Information Maximizing Generative Adversarial Network (InfoGAN), for synthesizing semantically rich images simultaneously conditioned on a pixel-level condition (PLC) and a macroscopic feature-level condition (FLC). Dimensionally, the PLC can only vary in the channel dimension from the synthesized image and is meant to be a task-specific input. The FLC is modeled as an attribute vector, $\boldsymbol{a}$, in the latent space of the generated image which controls the contributions of various characteristic attributes germane to the target distribution. During generation, $\boldsymbol{a}$ is sampled from $\mathbb{U}[0,1]$, while it is learned directly from the ground truth during training. An interpretation of $\boldsymbol{a}$ to systematically generate synthetic images by varying a chosen binary macroscopic feature is explored by training a linear binary classifier in the latent space. Experiments on a GPS trajectories dataset show that the proposed model can accurately generate various forms of spatio-temporal aggregates across different geographic locations while conditioned only on a raster representation of the road network. The primary intended application of the VAE-Info-cGAN is synthetic data (and label) generation for targeted data augmentation for computer vision-based modeling of problems relevant to geospatial analysis and remote sensing. 
\end{abstract}



\maketitle

\section{Introduction}\label{introduction}
Dearth of ``enough" and ``good-quality" labeled data for training supervised deep learning models for real-world applications of computer vision is an omnipresent problem that plagues the large-scale deployment of machine learning models in many domains. Even in situations where data is abundant, labels may be scarce, expensive, or difficult to obtain. While the subjective terms ``enough" and ``good-quality" have different definitions across different applications and domains, the term ``enough data" usually refers to an amount of data sufficient to avoid model over-fitting while the term ``good-quality data" usually refers to class-balanced, unbiased, and diverse datasets. Model over-fitting can lead to a lack of generalization while biased datasets usually lead to biased models. 

Creating a good quality, labeled dataset, which is large enough is difficult for many practical applications for a variety of reasons. Some of these reasons are: (i) collected data may be noisy and denoising/curation may be computationally or financially expensive, (ii) obtaining samples from rare classes/events may require large observation times which may be financially prohibitive, computationally expensive or infeasible, (iii) implicit biases in the collected data result in class imbalance or loss of diversity. There are many scenarios where the bias in the data is not a bug but a feature. For example, the geographic distribution of the volume of GPS data is expected to be proportional to population density. Such spatial biases can in turn be used as features informing computational models that help prioritize evacuations during natural disasters, monitor mobility to inform public health policies and epidemiological studies during epidemics, etc. However, in this paper, we are interested in situations where biased data is problematic and undesirable. Of specific interest in this paper are applications of computer vision for geospatial analysis and remote sensing --- a domain in which problems have often been cast as canonical computer vision tasks, e.g., land-cover classification cast as a multi-class semantic segmentation problem, pedestrian crosswalks detection cast as an object detection problem. 

Many geospatial and computational transportation applications involve modeling occurrence of infrequent events, e.g., identification and detection of (i) construction sites, (ii) road blocks, (iii) new roads, (iv) road closures, (v) junctions changing to roundabouts, (vi) addition or removal of pedestrian cross-walks, to name just a few. Some of these applications are critical to keep a geospatial mapping service up-to-date in real-time and can significantly improve the experience and safety of the users of a mapping service. For modeling such infrequent events, augmenting the training data is critical for training deep learning models that solve these tasks for four main reasons: (i) training data may be sparse in many geographical regions leading to poor diversity; (ii) events such as road constructions, road closures, etc., are associated with specific and rare temporal patterns, leading to a small dataset size; (iii) building a large enough labeled dataset requires a large observation time and may be prohibitively expensive or infeasible; (iv) satellite and/or sensor data being used may be stale in certain geographical regions due to latency or longer cadence of refresh. On the other hand, manual curation and annotation of these occurrences is expensive and not scalable.

In recent years, many approaches have been proposed to tackle the scarcity of labeled data specifically for computer vision tasks. These include unsupervised methods like consistency training \cite{xie2019unsupervised} and self-supervised learning, semi-supervised methods such as \cite{perez2019semi} which make intelligent use of unlabeled data, using generative models to create purely synthetic datasets \cite{8248284}, and active learning techniques \cite{gal2017deep} that gradually increase the number of labeled examples in an initially sparsely labeled dataset. Another approach is to perform automated data augmentation of the training dataset with synthetic data (and labels) to create a hybrid dataset consisting of real and synthetically generated examples such that supervised models can be trained using this hybrid dataset. The latter strategy requires access to a model that can conditionally generate meaningful samples from a target distribution in a controllable fashion. 

Towards this goal, in this paper, we propose a novel deep conditional generative model (DCGM) that simultaneously combines pixel-level conditions (PLC) and feature-level conditions (FLC), which are provided as inputs, to generate semantically rich synthetic images from a target distribution. We also propose a training methodology for the DCGM using pairs of PLC condition and the associated true sample from the target distribution obtained from different geographic locations. While designing the DGCM, emphasis was placed not just on generating high-quality and accurate synthetic images from a target distribution, but to expose to the user of the model, the ability to modulate the generated image either at the pixel-level (e.g., figure \ref{generated_heatmaps}) or at the macroscopic feature level (e.g., section \ref{latent_variable_tuning}, figure \ref{generated_tuning}) that are crucial in industrial and real-world applications (discussed in section \ref{motivation} and section \ref{appendix_temporal_changes} of the appendix) of generative models. Specifically, the model design facilitates generation of examples simply by modulating the PLC or FLC or both, so that targeted data augmentation of the training dataset can be performed. The evidence lower bound for the conditional VAE component of the model has also been derived from first principles. 

The performance of the proposed model is compared to 2 other variants of deep conditional generative models, namely, conditional Variational Autoencoder (cVAE) and conditional Generative Adversarial Network (cGAN). All three models are trained on a dataset described in detail in section \ref{data} for two different tasks. In one case, the models generate a single channel image while in the other, the models generate a 12-channel image. Quantitative comparison of the generated samples from all the 3 models with the ground truth for both tasks shows that the proposed DCGM outperforms the two other models.

A key consideration in designing the proposed DCGM is the fact that images are inherently \textit{multi-scale} --- information is encoded across a range of scales. An image can be represented as a 3D tensor since an image is an ordered collection of pixels with known channel-wise values at the microscopic level. However, an image is composed of lines, shapes, objects, contours, etc. at a macroscopic level. Simultaneously conditioning on the provided PLC (microscopic information) and the FLC (macroscopic information) takes advantage of the multi-scale nature of images. Furthermore, sampling from the desired regions of the target distribution is equivalent to learning the mapping from the PLC and FLC inputs to the target distribution. While the discussion in this paper targets geospatial applications, the modular architecture of the proposed DCGM is generic and can be used as a template for problems in other domains dealing with images or image-like data where diversifying the dataset with synthetically generated data is deemed useful.

The rest of this paper is organized as follows. Section \ref{motivation} provides a case study of a target geospatial application motivating the design of the proposed DCGM. Section \ref{previous} summarizes related work for generating synthetic data and manipulating the latent space of generative models. Section \ref{method} describes the detailed architecture of the proposed DCGM and the associated loss functions used to train the model. Section \ref{data} describes the preparation and particulars of the dataset on which the model is trained and the metrics used to evaluate trained models. Section \ref{exp} presents details of experiments while section \ref{results} summarizes the results from the experiments along with comparison of performance to other models. Section \ref{latent_variable_tuning} provides an example methodology of exploring the latent space of the model. Section \ref{discussion_and_conclusion} discusses the results and the model's limitations while providing some concluding remarks and directions for future work.

\section{Terminology and a Motivating Target Application}\label{motivation}
In this paper, \textit{traffic} refers to the systematic movement of vehicles and/or people across various motion modalities like walking, driving, etc. The \textit{traffic flow pattern} is defined as the characteristic distribution of traffic on a route as well as the interaction between travellers (such as pedestrians, drivers and their vehicles, etc.) and transport infrastructure (including highways, walkways, pedestrian crossings, road signage, traffic control devices like stop signs, traffic lights, etc.). In graph theory, road network corresponds to the graph, $\boldsymbol{G(V,E)}$, that encodes the topology, connectivity, and spatial structure of roads. Vertices, $\boldsymbol{V}$, denote intersections, end of roads or passages, or the starting/ending of road segments. Edges, $\boldsymbol{E}$, denote road segments between vertices of the graph. In this paper, we define the \textit{road network} to be simply a raster denoting the presence of a road. The \textit{binary road network} is a representation of the road network as an image where each pixel that has a road segment present is assigned the value 1 while pixels with no road segments present are assigned the value 0.

Traffic patterns can change due to seasonal variations (e.g., snowfall may cause certain roads to close down, traffic volume is usually higher in the day compared to the night), temporary road closures (e.g., due to accidents or road maintenance), changes in road infrastructure (e.g., new roads and detours replace older roads), addition or removal of traffic control devices (e.g., traffic lights or stop signs are added, a regular roundabout changes to a split roundabout), among other reasons. A key goal in computational transportation is to design systems that can monitor, understand, and interpret these changes in traffic patterns thereby inferring the underlying cause of the change. One approach to build such systems is to train supervised deep learning models to associate changes in traffic patterns to the cause of the change. Such systems can be used to reliably update and maintain a geospatial mapping service in an automated fashion without relying on voluntary information submitted by its users (whose information veracity or reporting certainty can be hard to determine) or on expensive manual curation which is not scalable. However, as discussed in section \ref{introduction}, changes like junctions changing to roundabouts, construction of new roads, etc. are infrequent events and obtaining enough training data may be prohibitively expensive in time and money. Generating synthetic training data reliably and inexpensively can ameliorate these problems and is the fundamental motivation behind the proposed DCGM. 

\begin{figure}
    \centering
    \includegraphics[width=0.48\textwidth]{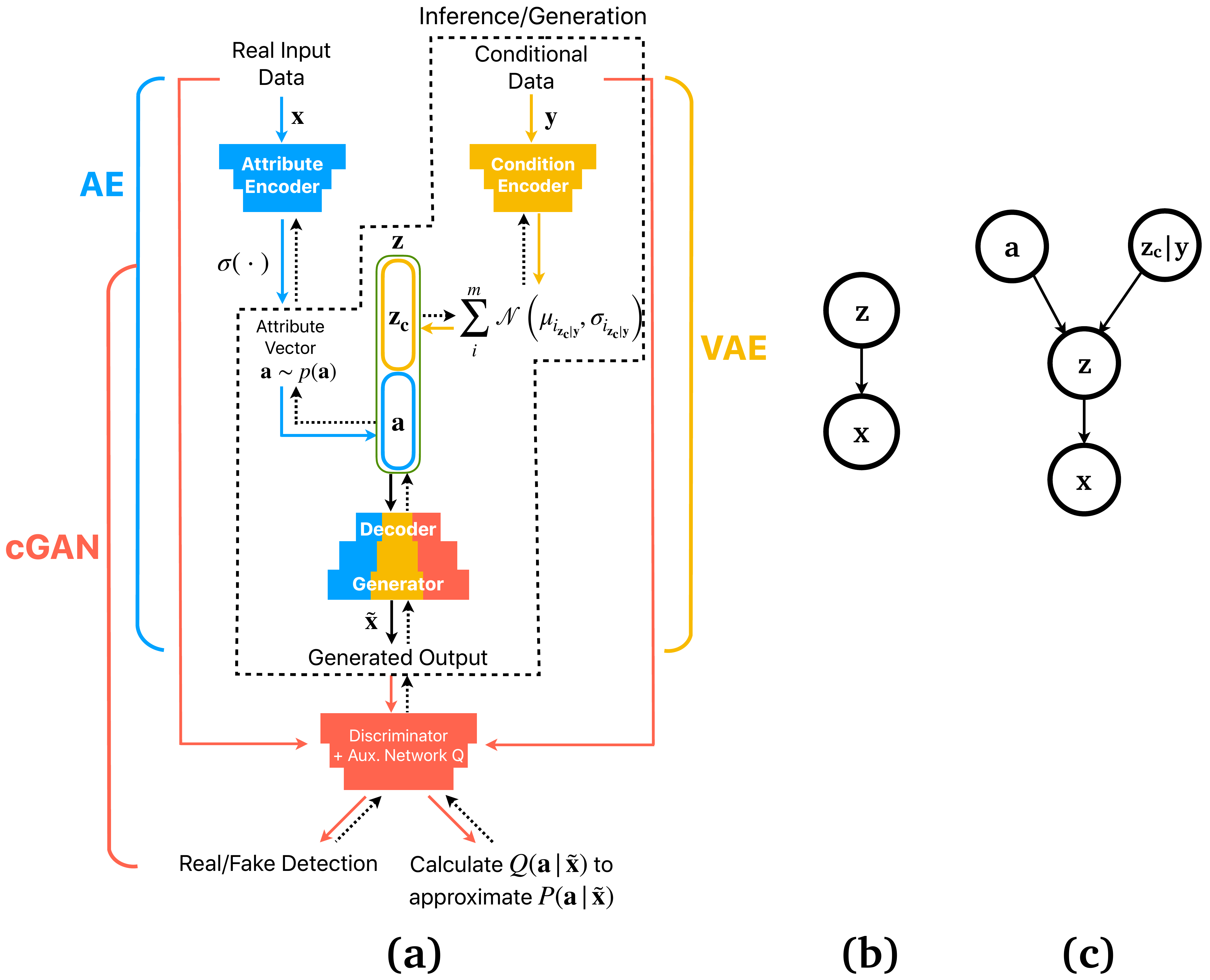}   
    \caption{(a) Schematic of the proposed DCGM architecture. Solid arrows show the flow of information in the forward pass while the dotted arrows denote the path for backpropagation during training. The black-dashed box encloses the parts active during inference. (b) The probabilistic graphical model (PGM) of a conventional VAE. (b) The PGM of the proposed DCGM.}
    \label{architecture}
\end{figure}

An example geospatial application for which synthetic data generated in a controllable fashion can be used to train a supervised deep learning model (SDLM) is detecting geographical locations where temporal changes in traffic patterns occur. Section \ref{appendix_temporal_changes} of the appendix discusses a methodology to train a SDLM which predicts locations where meaningful (not a simple difference of inputs at two different times) changes in traffic patterns occur using a synthetic training dataset generated by the proposed DCGM. In the discussion, details such as the nature of the inputs to the SDLM, the duration over which changes occur, neural architecture of the SDLM, etc. have been purposefully omitted to hone in on the key components of the methodology and restrain the discussion to a higher level of abstraction. While these details and the results from the discussed approach are the subject of a subsequent paper, the discussion highlights the motivations for some of the key ideas that influence the design of the proposed DCGM. Predictions from such a SDLM could help to update a map (or mapping service) with the latest road closures, newly built roads, etc. thereby ensuring freshness of the map and enhanced safety of its users.

\section{Related Work}\label{previous}
Two goals influence the design of the proposed DCGM: accurate generation of synthetic data guided by conditional inputs and manipulating the latent space to tune attributes of the generated data. Of the common genres of generative models (viz. autoregressive models, Variational Autoencoders (VAEs), normalizing flow models, Generative Adversarial Networks (GANs)) used in computer vision, VAEs have an accessible learned latent space which is interpretable \cite{kumar2017variational} while GANs are able to generate samples of higher visual quality \cite{style,biggan}. Synthetic data has been successfully generated using VAEs and GANs for a wide variety of data modalities including images \cite{betaVAE,vaegan,bowles2018gan,ganguli2019geogan}, music \cite{music}, text \cite{text}, etc. Methods for disentangling attributes in the latent space of VAEs and GANs have been proposed \cite{betaVAE,infogan} while modulating the latent space has been shown to help in controlling attributes in the generated data \cite{style,filtered,vaegan}.

Conditional versions of VAEs and GANs \cite{cvaes,cgans}, also called cVAE and cGAN, are of specific interest in this paper since they can generate samples, $\boldsymbol{\tilde{x}}$, that are close to the ground truth, $\boldsymbol{x} \sim p(\boldsymbol{x})$, from conditional inputs, $\boldsymbol{y} \sim p_c(\boldsymbol{y})$, and are therefore used as baselines to compare the results from the proposed model. The DCGM proposed in this paper is heavily inspired from the VAE-GAN model \cite{vaegan} where the generator of the GAN is also the decoder for the VAE. The VAE-GAN model \cite{vaegan} is however, not a conditional generative model. Furthermore, the proposed DCGM is trained using pairs of PLC inputs and the associated true probe aggregates at different geographic locations. This technique allows for incorporating into the learning objective a component that is aware of the quality of the generated output as compared to the ground truth, not only in terms of penalizing any difference between the distributions they represent, but also to penalize absolute differences in pixel values. One may analogize this to the content and style losses in neural style transfer or to the consistency loss in CycleGAN \cite{cyclegan}.

Access to the latent space in our model allows for developing systematic strategies (examples include \cite{interface,vaegan}) to manipulate the generated output by manipulating the latent representation of the output.  To ensure that there is strong correlation between the attributes and the generated samples, an auxiliary neural network (introduced in \cite{infogan}) is used to maximize the approximated mutual information between the attributes and the generated output. Similar models that semantically manipulate the latent space of generative models have also been proposed for other applications such as editing faces \cite{vaegan,interface}, motion transfer \cite{chan2019everybody}, and image-to-image translation \cite{ma2019novel}.

Nikolenko \cite{nikolenko2019synthetic} provides an extensive review of deep learning techniques for synthetic data generation primarily for computer vision tasks including a discussion of GAN-based approaches that can generate synthetic datasets with differential privacy guarantees. The primary target application of the proposed DCGM in this paper is data augmentation of existing real datasets with synthetically generated data so that the resulting hybrid dataset is class-balanced, diverse, and alleviates dataset bias. Applications across multiple domains have been shown to benefit from the use of synthetic data to tackle the lack of data diversity and \cite{nikolenko2019synthetic} provides many examples. These include medical imaging, robotics, virtual reality, physics-based models for video games, training computer vision models for self-driving cars, augmenting temporal sensor data, among others.

\section{Model Architecture and Training Methodology}\label{method}
A schematic of the different components of the proposed DCGM is shown in figure \ref{architecture}(a). In addition to describing the architecture of the model and the methodology used to train it, this section also introduces notation that will be used throughout the paper. Let $p\left( \boldsymbol{x} \right)$ denote the data distribution over a space of target outputs, $\boldsymbol{x} \in \mathcal{X}_{data}$. Let $\boldsymbol{\tilde{x}}$ denote the generated samples from the learned distribution $p(\boldsymbol{\tilde{x}})$. Here, $\boldsymbol{x}, \boldsymbol{\tilde{x}} \in \mathbb{R}^{h \times w \times c}$, where $h$, $w$, and $c$ denote the height, width, and number of channels in the images $\boldsymbol{x}$ and $\boldsymbol{\tilde{x}}$. Given the PLC input $ \boldsymbol{y} \in \mathcal{Y}$, $\boldsymbol{y} \in \mathbb{R}^{h \times w \times u}$ and FLC input $\boldsymbol{a} \in [0,1]^{d_a}$ of the target output, the goal is to learn and sample from the learned distribution, $p\left( \tilde{\boldsymbol{x}} \right)$, where $\tilde{\boldsymbol{x}} \in \mathcal{X}_{model}$, while minimizing the difference between $p\left( \boldsymbol{x} \right)$ and $p\left( \tilde{\boldsymbol{x}} \right)$. 

Towards this goal, we first learn $\boldsymbol{z_c} \in \mathbb{R}^{d_c}$, which is a compressed vector representation of the PLC input $\boldsymbol{y}$ using the encoder of a Variational Autoencoder (VAE). The embedding $\boldsymbol{z_c}$ produced from $\boldsymbol{y}$ is called the \textit{condition vector} while the FLC input $\boldsymbol{a}$ is called the \textit{attribute vector}. Noting that $\boldsymbol{a}$ is a $d_a$-dimensional vector (with all entries being real numbers between 0 and 1) and $\boldsymbol{z_c}$ is a $d_c$-dimensional vector, $\boldsymbol{z_c}$ is concatenated with $\boldsymbol{a}$ to form the latent variable $\boldsymbol{z} \in \mathbb{R}^{d_a + d_c}$, which is input to the generator ($G$) of the Generative Adversarial Network (GAN) component (also, decoder of VAE component) shown in figure \ref{architecture}(a). In the proposed model, the concatenation of $\boldsymbol{z_c}$ and $\boldsymbol{a}$ produces the $(d_a + d_c)$-dimensional vector $\boldsymbol{z}$, which is the latent variable that encodes information of the PLC and FLC. Another approach, not explored in this paper, could be to learn (via a neural network) the optimal combination of $\boldsymbol{z_c}$ and $\boldsymbol{a}$ to produce $\boldsymbol{z}$. $\boldsymbol{a}$ is modeled as $\boldsymbol{a} \sim \mathbb{U}[0,1]$ and is learned using an encoder, with sigmoid activation in the last layer, from the real data $\boldsymbol{x}$ during training. $\boldsymbol{z_c}$ and $\boldsymbol{a}$ are both continuous vectors. 

The latent variable, $\boldsymbol{z}$, is up-sampled by the generator to generate the output. The encoder of the VAE, the attribute encoder, and the generator have fully convolutional architectures with additional design choices described in section \ref{exp}. The GAN is conditioned on $\boldsymbol{y}$ via $\boldsymbol{z_c}$ and the FLC via $\boldsymbol{a}$. The discriminator ($D$) of the GAN is trained to distinguish the real distribution $P(\boldsymbol{x}, \boldsymbol{y})$ from the generated distribution $P_{\boldsymbol{a}, \boldsymbol{z_c}}(\tilde{\boldsymbol{x}}, \boldsymbol{y})$. The final layer of the discriminator is connected to an auxiliary fully connected neural network $Q$ that outputs parameters for $Q(\boldsymbol{a} | \tilde{\boldsymbol{x}})$ for estimating the true but intractable posterior $p(\boldsymbol{a} | \tilde{\boldsymbol{x}})$ which is used to maximize the mutual information between $\boldsymbol{a}$ and the generated image $\tilde{\boldsymbol{x}}$. Thus, in addition to being a binary classifier, the discriminator of the GAN also provides incentive for a larger influence of the desired attributes (provided via $\boldsymbol{a}$) on the generated data distribution. Based on the architecture of the model, we name the proposed DCGM as VAE-Info-cGAN. 

The model is trained end-to-end with corresponding ($\boldsymbol{x},\boldsymbol{y}$) pairs from different geographic locations to generate $\tilde{\boldsymbol{x}}$ close to $\boldsymbol{x}$. Once the model has been trained, only $\boldsymbol{y}$ is required at generation time while $\boldsymbol{a}$ is sampled from $\mathbb{U}[0,1]$. Since there is no upper bound on the pixel values across the channels of the images, the images are normalized to meaningfully train the neural network and avoid spurious effects from the large dynamic range of the images. During training, the input, $\boldsymbol{x}$, to the attribute encoder is log-normalized using $\boldsymbol{x'} = \log(1 + \boldsymbol{x})$ thereby training the model to generate log-normalized images, $\boldsymbol{\tilde{x}'}$. To obtain the final generated image, $\boldsymbol{\tilde{x}}$, an inverse transform, $\boldsymbol{\tilde{x}} = \exp({\boldsymbol{\tilde{x}'}})-1$, is applied to $\boldsymbol{\tilde{x}'}$. Each of the components of the model's architecture are explained in further detail in the sub-sections that follow.

\textbf{Attribute Encoder (AE):} The aim of AE is to learn a compressed encoding for the real input data and is trained similar to a plain autoencoder. This component contributes equation \ref{Lae} to the total loss function of the model which is the pixel-wise mean squared error between the real and generated data, providing incentive for pixel-wise similarity between the two. 
\begin{equation} \label{Lae}
L_{AE} = MSE(x, \tilde{x}) 
\end{equation}

\textbf{Variational Autoencoder (VAE):} A conventional VAE tries to generate samples from a given data distribution by learning a parameterization $p_{\theta} \left( \boldsymbol{x} \right) = \int{p(\boldsymbol{z}) p_{\theta} \left( \boldsymbol{x}|\boldsymbol{z} \right)} d\boldsymbol{z}$ of the true data distribution $p(\boldsymbol{x})$, where $p(\boldsymbol{z})$ is the prior of the latent variable, $\boldsymbol{z}$, and can be described as following the probabilistic graphical model in figure \ref{architecture}(b). The intractibility of $p_{\theta}\left( \boldsymbol{x} \right)$ is tackled using amortized inference with a neural encoder that learns the parameters, $\phi$, of the variational posterior, $q_{\phi}(\boldsymbol{z}|\boldsymbol{x})$, that approximates the true posterior, $p(\boldsymbol{z}|\boldsymbol{x})$, directly from $\boldsymbol{x}$ by maximizing the associated evidence lower bound (ELBo) of the marginal likelihood, $p_{\theta}(\boldsymbol{x})$. In the conditional setting of the proposed DCGM demonstrated using the probabilistic graphical model shown in figure \ref{architecture}(c), the ELBo is slightly different and is derived below. The design of the model assumes that the two conditional inputs, PLC and FLC, are independent. For the variables $\boldsymbol{a}$ and $\boldsymbol{z_c}$, $p(\boldsymbol{a}) \sim \mathbb{U}[0,1]$ and $p(\boldsymbol{z_c} | \boldsymbol{y})$ is modeled as a mixture of $m$ Gaussians. The prior $p_{\theta}(\boldsymbol{z})$ is modeled as $\mathcal{N}(0,I)$. Denoting $\boldsymbol{z_c}|\boldsymbol{y}$ as the variable $\boldsymbol{w}$, we then observe that 
\begin{equation}
    p_{\theta}(\boldsymbol{x}) = \sum_{\boldsymbol{z}} q_\phi(\boldsymbol{z}\big|\boldsymbol{a},\boldsymbol{w}) \frac{p_{\theta}(\boldsymbol{x},\boldsymbol{z})}{q_\phi(\boldsymbol{z}\big|\boldsymbol{a},\boldsymbol{w})}
\end{equation}
Using Jensen's inequality yields 
\begin{equation}
    \log p_{\theta}(\boldsymbol{x}) \geq \sum_{\boldsymbol{z}} q_\phi(\boldsymbol{z}\big|\boldsymbol{a},\boldsymbol{w}) \log\left(\frac{p_{\theta}(\boldsymbol{x},\boldsymbol{z})}{q_\phi(\boldsymbol{z}\big|\boldsymbol{a},\boldsymbol{w})}\right)
\end{equation}
Simplifying the right hand side of the above inequality yields the ELBo on the marginal log-likelihood of $p_\theta(\boldsymbol{x})$ shown in equation \ref{Lvae3}, which is used to train the VAE component of the proposed DCGM.
\begin{equation} \label{Lvae3}
L_{VAE} = D_{KL} \left[q_{\phi}(\boldsymbol{z} | \boldsymbol{a}, \boldsymbol{w}) \| p_\theta(\boldsymbol{z})\right] - \mathbb{E}_{q_{\phi}(\boldsymbol{z}|\boldsymbol{a},\boldsymbol{w})} \left[\log p_{\theta}(\boldsymbol{x}|\boldsymbol{z})\right] 
\end{equation}
The pixel values in images in the target distribution of the applications described in this paper are positive whole numbers. Since the model learns to generate log-normalized images, the distribution $p_{\theta}\left( \boldsymbol{x}|\boldsymbol{z} \right)$ is modeled as a log-normal distribution. Given that the encoder of the VAE produces a compressed representation of the input PLC, the encoder of the VAE is also called the condition encoder (CE).

\textbf{Information Maximizing Conditional GAN (Info-cGAN):} In a typical GAN, the generator is responsible for generating samples while the discriminator evaluates and distinguishes the generated data from real data. A conditional GAN takes conditional data as part of the input to guide generation. The proposed combination of VAE and GAN allows us to learn a mapping from the distribution of conditional inputs, $(\boldsymbol{a}, \boldsymbol{y})$, to the target distribution, $p(\boldsymbol{x})$. Similar to InfoGAN \cite{infogan}, an additional neural network, $Q$, is added at the end of the discriminator to ensure high correlation between the attributes and the generated data. The discriminator outputs not only the classification of real and generated samples, but also parameters of the conditional distribution $Q(\boldsymbol{a}|\tilde{\boldsymbol{x}})$ (which is modeled as a Gaussian). The variational lower bound of the mutual information between $\boldsymbol{a}$ and $\tilde{\boldsymbol{x}}$ is approximated using this auxiliary neural network and maximized. 

\textbf{Loss Function and Training Methodology:} The input to the discriminator, $D$, whether it be $\boldsymbol{x}$ or $\tilde{\boldsymbol{x}} = G(\boldsymbol{z_c}, \boldsymbol{a})$, is concatenated with the PLC, $\boldsymbol{y}$, in the channel dimension. Let $D(\cdot)$ denote the discriminator's estimate of the probability that the input to the discriminator is real. Denoting the concatenated versions of $\boldsymbol{x}$ and $\tilde{\boldsymbol{x}}$ as $\boldsymbol{X}$ and $\tilde{\boldsymbol{X}}$, respectively, the loss functions (non-saturating loss is used for the GAN) for the discriminator (eqn. \ref{Ldisc}), the generator (eqn. \ref{Lgen}), and the information loss (eqn. \ref{Linfo}) are 
\begin{align}
    L_{disc} &= - \mathbb{E}_{p_{\text{data}}} \left[\log D(\boldsymbol{X}) \right] - \mathbb{E}_{\boldsymbol{z_c},\boldsymbol{a}} [\log (1 - D(\tilde{\boldsymbol{X}}))] \label{Ldisc},\\
    L_{gen} &= - \mathbb{E}_{\boldsymbol{z_c},\boldsymbol{a}} [\log (D(\tilde{\boldsymbol{X}}))] \label{Lgen}, \\
    L_{info} &= - \mathbb{E}_{\boldsymbol{a} \sim p(\boldsymbol{a}), \tilde{\boldsymbol{x}} \sim G(\boldsymbol{z_c},\boldsymbol{a})} \left[ \log Q(\boldsymbol{a}|\tilde{\boldsymbol{x}}) \right] \label{Linfo}.
\end{align}
The proposed DCGM is optimized by alternately minimizing the total discriminator loss (eqn. \ref{eqD}) and generator loss (eqn. \ref{eqG}), which are the weighted sums of the losses of all four components. 
\begin{align}
    L_{D_{total}} = \lambda_1 L_{AE} + \lambda_2 L_{VAE} + \lambda_3 L_{disc} + \lambda_4 L_{info} \label{eqD}.\\
    L_{G_{total}} = \lambda_1 L_{AE} + \lambda_2 L_{VAE} + \lambda_3 L_{gen} + \lambda_4 L_{info} \label{eqG}.
\end{align}
Here, $\lambda_1, \lambda_2, \lambda_3$, and $\lambda_4$ are hyperparameters set by the user of the DCGM and can be used to control the influence of the four different components of the model on the generated samples.

\section{Dataset Preparation and Evaluation Metrics}\label{data}
In addition to satellite imagery and sensor data, a variety of proprietary datasets serve as rich sources of geospatial data. One such proprietary dataset is \textit{probe data} \cite{tcrunch}, which is a privacy-preserving, structured, and sequential dataset\footnote{Probe data is similar to GPS trace trajectories found in publicly available datasets such as \cite{msrtaxi,ucigpsdataset}. More details of probe data can be found in the section titled ``Probe data and privacy'' in \cite{tcrunch}.}. Such a sequential dataset can be pre-processed and transformed to semantically meaningful, multi-channel, \textit{image-like} data (raster images) making them amenable for use by downstream computer vision models that solve various geospatial tasks without using satellite imagery. 

Two examples of such transformations on probe data are proposed in this paper. They are (i) count-based raster maps (CRM) and (ii) bucketed heading count-based raster maps (HCRM). A count-based raster map is an array of equally sized cells where the value of each cell is a positive whole number that represents the number of occurrences of a given quantity over a fixed time period in that cell. Probe data encodes multiple motion-modalities like driving, walking, etc. For the experiments in this paper, probe data for a given representative time interval, $\Delta t$, is filtered using a pre-trained, multi-layer LSTM-based, motion-modality filter to extract the data corresponding to driving modality. CRM is the count-based raster map obtained from the filtered probe data on a grid whose cells correspond to zoom-24 tiles\footnote{A tile resulting from viewing the spherical Mercator projection coordinate system (EPSG:3857) \cite{epsg_3857} of earth as a $2^{24}$ $\times$ $2^{24}$ grid. This corresponds to a spatial resolution of approximately 2.38 m at the equator.} on the surface of the earth. If the count-based raster map is formed by bucketing based on the direction in which the probe is heading into 12 buckets of 30\degree, and each bucket is represented as a channel in the image, we obtain HCRM. HCRM directly encodes the directionality of probe data. Summing across all the 12 channels of HCRM at each pixel recovers the CRM.

In spectral imaging, which is used extensively for remote sensing applications, different spectral bands (represented as channels) of the image of a geographic location allow extraction of additional information compared to their RGB (visible spectrum) counterparts. Analogously, HCRM is one of many ways probe data can be represented as multi-channel images which encode additional information that is useful for many downstream computer vision-based applications. CRM and HCRM are based on pixel-wise counts which are non-negative integers with no upper bound. The amount of traffic in the chosen geographical region and the observation time interval, $\Delta t$, determine the magnitude of the pixel values across channels. 

Two versions of the proposed DCGM architecture are trained to synthetically generate CRM and HCRM, respectively, such that the PLC input is only the binary road network (which is a single channel input, i.e. $u=1$). While a wide array of macro-level features of the synthesized image, supplied via the FLC input, are of interest (e.g., noise level of probes, noise in adherence of the probe to the PLC, time duration of observations\footnote{One may intuitively expect probe density to scale linearly with size of time interval $\Delta t$, but seasonal and other nonlinear effects influence this density, which the FLC may be tasked to learn.} ($\Delta t$), etc.), section \ref{latent_variable_tuning} explores a methodology for tuning the latent space of the trained model such that varying a single scalar parameter is equivalent to changing $\Delta t$.

A pixel for images considered in this paper is a zoom-24 tile. The ground truth and generated images have a resolution of a zoom-17 tile ($128 \times 128$ pixels) so that $h=w=128$. Two 70-20-10 training-validation-testing datasets are prepared for CRM and HCRM, respectively. In general, evaluating a generative model is a challenging problem and a wide array of techniques are used to evaluate them \cite{theis2015note}. However, for the evaluation of the models trained to generate CRM and HCRM, the ground truth is available in the training, validation, and test sets. The attribute vector is calibrated on a validation set in a geographical region similar to the test set by regressing on the ground truth. This attribute vector can then be used in addition to road networks in the test set to generate images which can be compared with the ground truth images, both visually and quantitatively. 

Analogous to models built to reconstruct data, a metric that can be used to evaluate conditional generative models of the type proposed in this paper, where the ground truth is available in the test set, is peak signal-to-noise ratio (PSNR). Since there is no bound on pixel values across channels in CRM or HCRM, a maximum possible pixel value of the image cannot be defined. Instead, the quantitative metric most relevant to our application is the normalized root-mean-squared error (RMSE) of the pixel-wise difference between the generated image and the ground truth image. We define average percentage normalized deviation (APND) as
\begin{equation*}
    APND = \frac{1}{|\mathcal{D}_{\text{test}}|} \sum_{\boldsymbol{x} \in \mathcal{D}_{\text{test}}} \left[ \frac{\Vert \boldsymbol{x} - \boldsymbol{\tilde{x}}\Vert_2}{\Vert \boldsymbol{x} \Vert_2} \times 100 \right]
\end{equation*}
where, $\mathcal{D}_{\text{test}}$ denotes the test set, $|\mathcal{D}_{\text{test}}|$ is the number of examples in the test set, $\Vert \boldsymbol{x} - \boldsymbol{\tilde{x}}\Vert_2^2 = \sum_i^h \sum_j^w \sum_k^c (\boldsymbol{x}(i,j,k) - \boldsymbol{\tilde{x}}(i,j,k))^2$, and $\Vert \boldsymbol{x} \Vert_2^2 = \sum_i^h \sum_j^w \sum_k^c \boldsymbol{x}(i,j,k)^2$. APND is the metric used to compare trained models in this paper. Additionally, visual inspection of generated images is used for qualitative comparison of trained models.

\section{Experiments}\label{exp}
This section summarizes observations and conclusions from experiments conducted on the proposed DCGM's architecture and training methodologies with the goal of obtaining high quality generated samples. The observations are itemized in a component-wise fashion and are presented for both the use-cases, namely, synthetic generation of CRM and HCRM. 

\textbf{(i) Attribute Encoder (AE):} Experiments with the depth of this component yielded significantly better results when the encoder was deeper for both CRM and HCRM generation. We varied the depth from 7 convolutional layers and an inception layer to 15 convolutional layers followed by an inception layer \cite{szegedy2017inception}. The deeper version showed an approximately 12\% reduction in the total training loss, accompanied with sharper visual quality of the generated samples. The inception layer, which allows the encoder to capture features with different sizes of receptive fields positively influences the quality of the generated sample. Adding dilated convolutions in the encoder was also helpful which effectively increases the receptive field of the layers without increasing the number of parameters to be trained. For the generation of HCRM specifically, converting all convolutional layers from 2D-convolutions to 3D-convolutions helps to more effectively learn the correlations in the channel dimension. Quantitatively, using 3D-convolutions resulted in the reduction of the total training loss by approximately 7\% compared to the case when 2D-convolutions were used when the model was tasked to synthetically generate HCRM. 

\textbf{(ii) Condition Encoder (CE):} Adding inception layers to the VAE encoder proved to increase the quality of the generated samples for generating both CRM and HCRM. Furthermore, adding skip connections between the VAE encoder and VAE decoder (also the generator for the GAN) was very effective and improved the visual quality and sharpness of the generated images. Quantitatively, presence of skip connections translated to a reduction of 8\% in the the total loss and a 10\% reduction in the APND in the test set for both CRM and HCRM generation compared to the models without skip connections. Skip connections ensure that the generator receives both local and global features from the condition $\boldsymbol{y}$ (in this case, the binary road network). Modeling $p(\boldsymbol{z})$ as a mixture of $m$ Gaussians as opposed to a single Gaussian increased the visual quality of the generated samples and also proved to be helpful in minimizing the KL-divergence component in the ELBo. Quantitatively, $m=20$ resulted in a 5\% reduction in the total training loss for both CRM and HCRM generation compared to $m=1$. The covariance matrix of the Gaussians, $\{\sigma_{i_{\boldsymbol{z_c}}|\boldsymbol{y}}\}_{i=1}^{m}$, are modeled as diagonal matrices and the condition encoder learns the diagonal values in addition to the mean. Modeling the multivariate Gaussians as heteroscedastic as opposed to homoscedastic ($\sigma_{i_{\boldsymbol{z_c}}|\boldsymbol{y}} = \zeta_i^2 \boldsymbol{I}$, where $\zeta_i$ is a scalar and $\boldsymbol{I}$ is the identity matrix) yielded visually sharper generated samples.  

\textbf{(iii) Generator (G):} The input to the generator is the concatenation of both $\boldsymbol{a}$ and $\boldsymbol{z_c}$. The lengths of both these vectors was experimented with. Conforming with intuition, the smaller the length of $\boldsymbol{a}$, the more changes we can impose on the generated image by tuning a single element in the attribute vector. The generator is primarily composed of up-sampling layers that takes the input of dimension $1\times1\times(d_a+d_c)$ and eventually outputs generated data of dimension $128\times128\times c$, where $c=1$ for CRM and $c=12$ for HCRM. Increasing the depth of the generator (deepest being 7 alternating convolutional and upsampling layers) produces outputs with higher contrast. In early experiments with transposed convolutional layers, the generated images presented glaring checkerboard patterns which is alleviated by replacing the transposed convolutions with nearest neighbor interpolations (resize layer) together with a convolutional layer \cite{odena2016deconvolution} that preserves the input dimensions. 

\textbf{(iv) Discriminator (D):} The discriminator consists of 6 convolutional layers and an inception module. The penultimate layer of the discriminator bifurcates into two branches. One of the branches is a convolutional layer that produces the logit corresponding to the estimate of the probability that the input to the discriminator is real. The other branch consists of two fully connected layers and is the auxiliary neural network $Q$ (that approximates the true posterior $P(\boldsymbol{a}|\tilde{\boldsymbol{x}})$ modeling it as a Gaussian $Q(\boldsymbol{a}|\tilde{\boldsymbol{x}})$) that outputs the mean and variance of the Gaussian $Q(\boldsymbol{a}|\tilde{\boldsymbol{x}})$. 

\textbf{(v) Ablation Study:} An architectural ablation study was performed where the discriminator (and the auxiliary network $Q$) was removed and the hyperparameters $\lambda_3$ and $\lambda_4$ set to zero. This significantly downgraded the visual quality of the generated samples which also manifested as a 30\% higher APND in the test set compared to the model with the discriminator present. Additionally, comparing the absolute values of $L_{AE}$ and $L_{VAE}$ between the models with and without the discriminator (and the auxiliary network $Q$), the presence of the discriminator resulted in approximately 5-15\% reduction in these absolute values across the various architectural designs and depths of the components that was experimented with. This suggests that the discriminator, which evaluates the high-level features of its input and thereby provides a holistic evaluation of the generated samples, helps the model train better. This signal, in addition to the auxiliary network $Q$ incentivising a stronger correlation between $\boldsymbol{a}$ and the generated samples, also helps the other components of the model to contribute more effectively so that the model as a whole learns to generate good quality samples.   

Based on these experiments, the neural architectures of the components of the models that yield the best performance is described in detail in section \ref{appendix_neural_architecture} of the appendix.

\section{Results}\label{results}
As mentioned in section \ref{data}, the APND metric (along with a 95\% confidence interval) is used to select the best performing models by evaluating them on the validation set. Results of this metric are reported on the test set. For minimizing the total discriminator and generator losses, the Adam optimizer with $\beta_1 = 0.9$, $\beta_2 = 0.999$, $\epsilon = 1 \times 10^{-8}$, and an initial learning rate of 1 x $10^{-3}$ is used. Decaying the initial learning rate exponentially with an exponent of 0.95 in a stair-case (length being 1/5 of an epoch) fashion was found to be useful for loss convergence. We found that the best performance was obtained when $\lambda_1 = \lambda_2 = \lambda_3 = \lambda_4 = 1 $ in equations \ref{eqD} and \ref{eqG}. Limiting error signals to relevant networks suggested by \cite{vaegan} did not provide significant improvement in the generated results. 

The performance of the proposed model is compared to that of cVAE and cGAN. To ensure fair comparison, two versions of the cVAE and cGAN are trained. The first version only uses the PLC (binary road network) as the input. In the second version, an AE component is added to the cVAE and cGAN to provide the FLC, which is trained using an additional MSE loss analogous to VAE-Info-cGAN. Furthermore, the neural architectures of the encoder of the cVAE, decoder of the cVAE, generator of cGAN, and discriminator of the cGAN are same as that of the CE, G, G, and D components of the VAE-Info-cGAN, respectively. In the cGAN, the CE (same architecture as VAE-Info-cGAN) is used to compress the PLC into the vector input fed to the generator. For the variants of cGAN and cVAE models with the AE present, the FLC is concatenated to the encoding of the PLC before it is fed as input to the generator of the cGAN while, the FLC is concatenated in the latent space of the cVAE to the encoding of the PLC and then modeled as the latent vector which is fed to the decoder of the cVAE. 

\begin{figure*}
    \centering
    \begin{subfigure}{0.28\textwidth}
        \centering
        \includegraphics[width=0.83\textwidth]{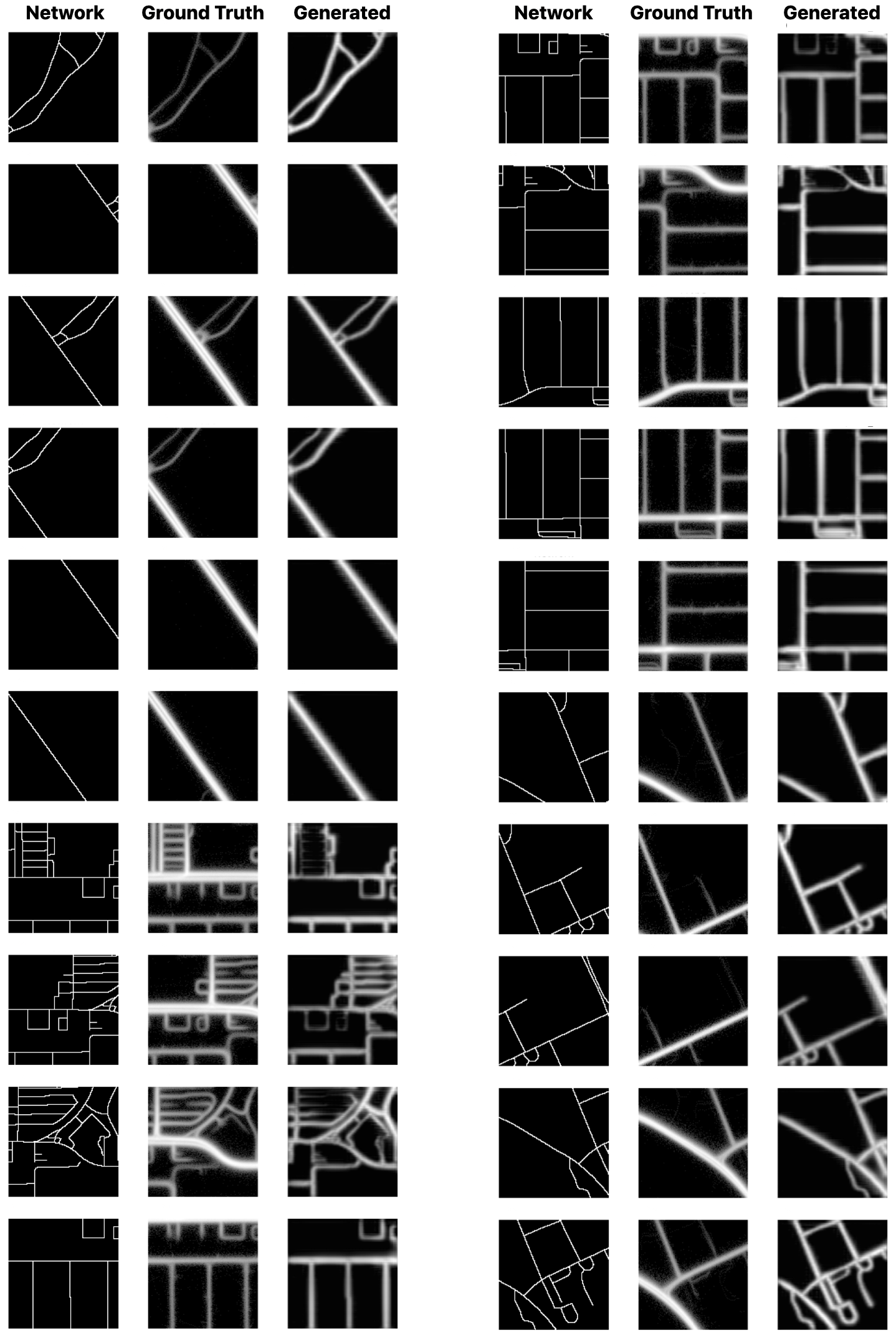}
        \caption{Generated samples of CRM}
        \label{density_heatmaps}
    \end{subfigure}
    \hfill
    \begin{subfigure}{0.49\textwidth}
        \centering
        \includegraphics[width=\textwidth]{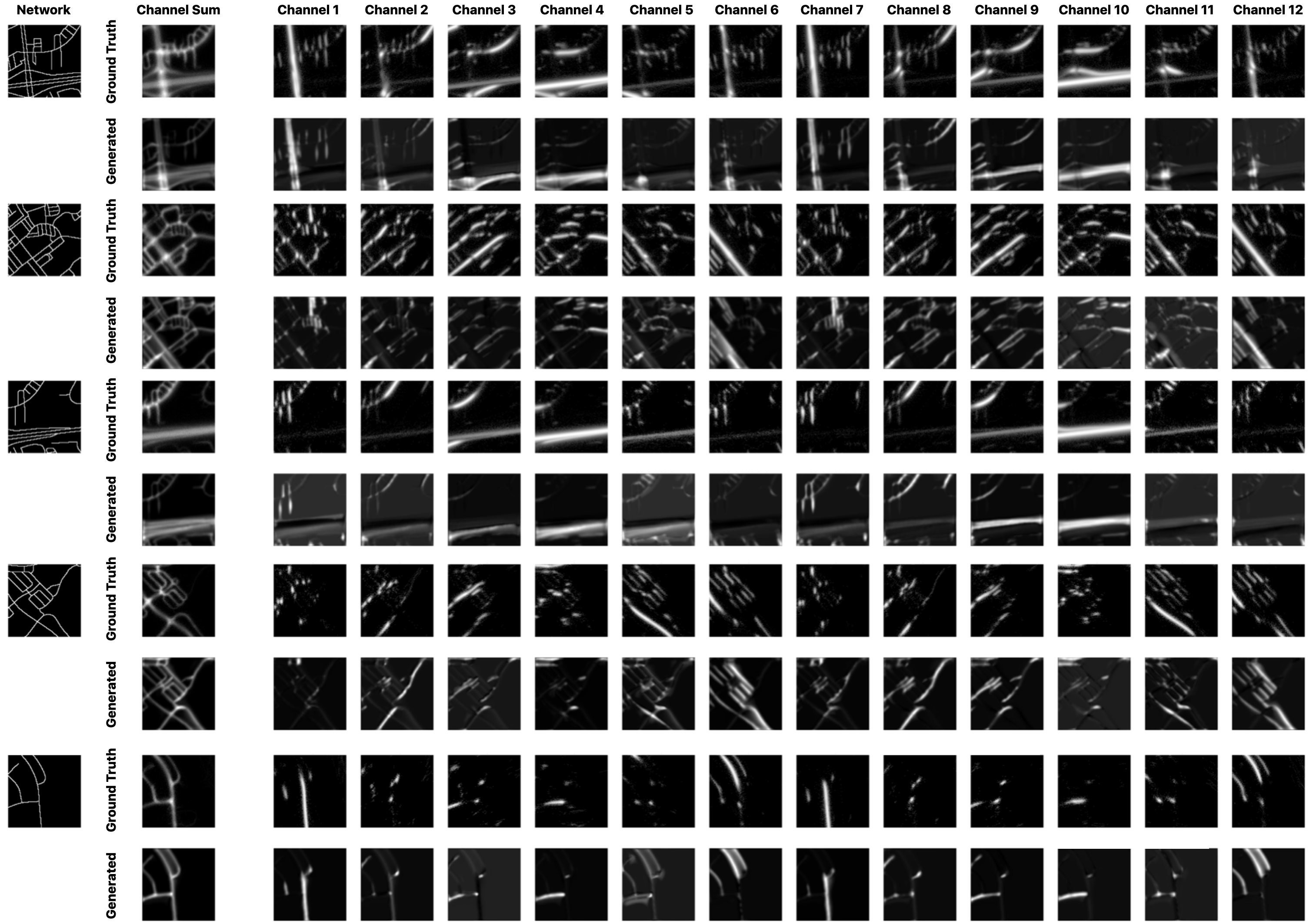}
        \caption{Generated samples of HCRM}
        \label{heading_heatmaps}
    \end{subfigure}
    \caption{(a) Comparison of ground truth and generated samples for 20 examples of road network (left), ground truth (middle), and generated CRM (right). (b) Comparison of ground truth and generated samples for 5 examples of road network (left), ground truth (12 channels top), and generated HCRM (12 channels bottom). The sum of the 12 channels in HCRM, which corresponds to the CRM, is also shown.}
    \label{generated_heatmaps}
\end{figure*}

Sample outputs from the best trained VAE-Info-cGAN models for generating CRM and for generating HCRM are shown in figures \ref{density_heatmaps} and \ref{heading_heatmaps} respectively. In each case, the input condition (binary road network) and the ground truth image are also provided side-by-side for comparison. Quantitative comparison with both variants of the cVAE and cGAN are shown in table \ref{results_table} for generating both CRM and HCRM. VAE-Info-cGAN is seen to perform better than its other conditional generative model counterparts. The VAE-Info-cGAN can be viewed as a model that incorporates the best of both the cVAE and cGAN models with additional modifications.

\begin{table}[hbt!]
    \caption{Comparison of Models (Lower APND is better) \label{results_table}}
    \centering
    \begin{tabular}{lcc}
        \toprule
        \centering{Model}      &  APND on CRM                &  APND on HCRM              \\
        \midrule
        cVAE (only PLC)        &  $1.37 \pm 0.02$\%          & $1.84 \pm 0.03$\%          \\
        cVAE (PLC and FLC)     &  $1.29 \pm 0.02$\%          & $1.67 \pm 0.03$\%          \\
        cGAN (only PLC)        &  $1.03 \pm 0.02$\%          & $1.39 \pm 0.03$\%          \\ 
        cGAN (PLC and FLC)     &  $0.83 \pm 0.02$\%          & $1.23 \pm 0.03$\%          \\
        \textbf{VAE-Info-cGAN} &  $\mathbf{0.53 \pm 0.02\%}$ & $\mathbf{0.98 \pm 0.03\%}$ \\
        \bottomrule
    \end{tabular}
\end{table}

\section{Tuning of Latent Representation}\label{latent_variable_tuning}
The VAE-Info-cGAN is designed to have an accessible latent space. This latent space is composed of both the attribute vector, $\boldsymbol{a}$, and the condition vector, $\boldsymbol{z_c}$, which are concatenated to form the latent representation, $\boldsymbol{z}$. One example approach to interpreting and manipulating the latent space to generate samples in a controlled fashion by varying a desired macro-level feature is presented in this section. The macro-feature chosen is the time duration of the observed probe data, $\Delta t$. This is an important feature since it is often important to know how the magnitude of data varies at a given fixed location (i.e. fixed binary road network, $\boldsymbol{y}$) as the observation time interval, $\Delta t$, varies. A proxy metric for this is the sum of all pixel values in the image. 

Shen et al. \cite{interface} show that attributes of generated images can be manipulated directly by varying the latent vector based on the corresponding linear subspace of the latent space of an unconditional GAN assuming that for any binary attribute, there exists a hyperplane in the latent space of the GAN that serves as the decision boundary. Extending this idea to the proposed DCGM, with the caveat that the latent representation, $\boldsymbol{z_c}$, of the VAE is fixed (since this is obtained from the provided binary road network), we assume that the attribute vector, $\boldsymbol{a}$, encodes multiple features including the binary macro-level feature associated with probe-density. By varying $\boldsymbol{z}$ in the direction of the normal, $\boldsymbol{n}$, of the decision hyperplane of this binary feature, it is shown that the pixel magnitude of the generated samples can be systematically controlled without re-training the DCGM. 

For a given input condition, $\boldsymbol{y}$ (thereby fixing $\boldsymbol{z_c}$), the trained DCGM is used to generate 10,000 images by randomly sampling $\boldsymbol{a}$ from $\mathbb{U}[0,1]$. For each of the generated images, the summation over all pixels is calculated and used as a proxy metric for the data observation interval. The mean of the pixel sum over all the generated images, $\mu_{PS}$, is used as the threshold for automatically labeling generated images with a higher sum than $\mu_{PS}$ as +1 and those with a lower sum than $\mu_{PS}$ as -1. A linear SVM is trained on the FLC vector, $\boldsymbol{a}$, of the generated images with labels as described. 90\% of the data is used for training this SVM while 10\% of the data is used for validation. 

This methodology was tested on 5000 different, randomly chosen examples from the test set. The accuracy of the trained SVM for all the 5000 examples exceeded 95\%. The decision hyperplane for the SVM and the unit normal vector $\boldsymbol{n}$ to this hyperplane can be calculated in closed form. We pick $\boldsymbol{a}_{boundary}$ closest to the decision boundary and vary it with different $\alpha$ such that $\boldsymbol{a}_{edit} = \boldsymbol{a}_{boundary} + \alpha \boldsymbol{n}$. Concatenating $\boldsymbol{a}_{edit}$ and $\boldsymbol{z_c}$ to produce latent representations $\boldsymbol{z}$ that vary in the direction of the normal to the decision hyperplane, we use the resulting $\boldsymbol{z}$-vectors to generate images with increasing ($\alpha > 0$) or decreasing ($\alpha < 0$) probe density. 

\begin{figure*}
    \centering
    \begin{subfigure}{0.25\textwidth}
        \centering
        \includegraphics[width=0.76\textwidth]{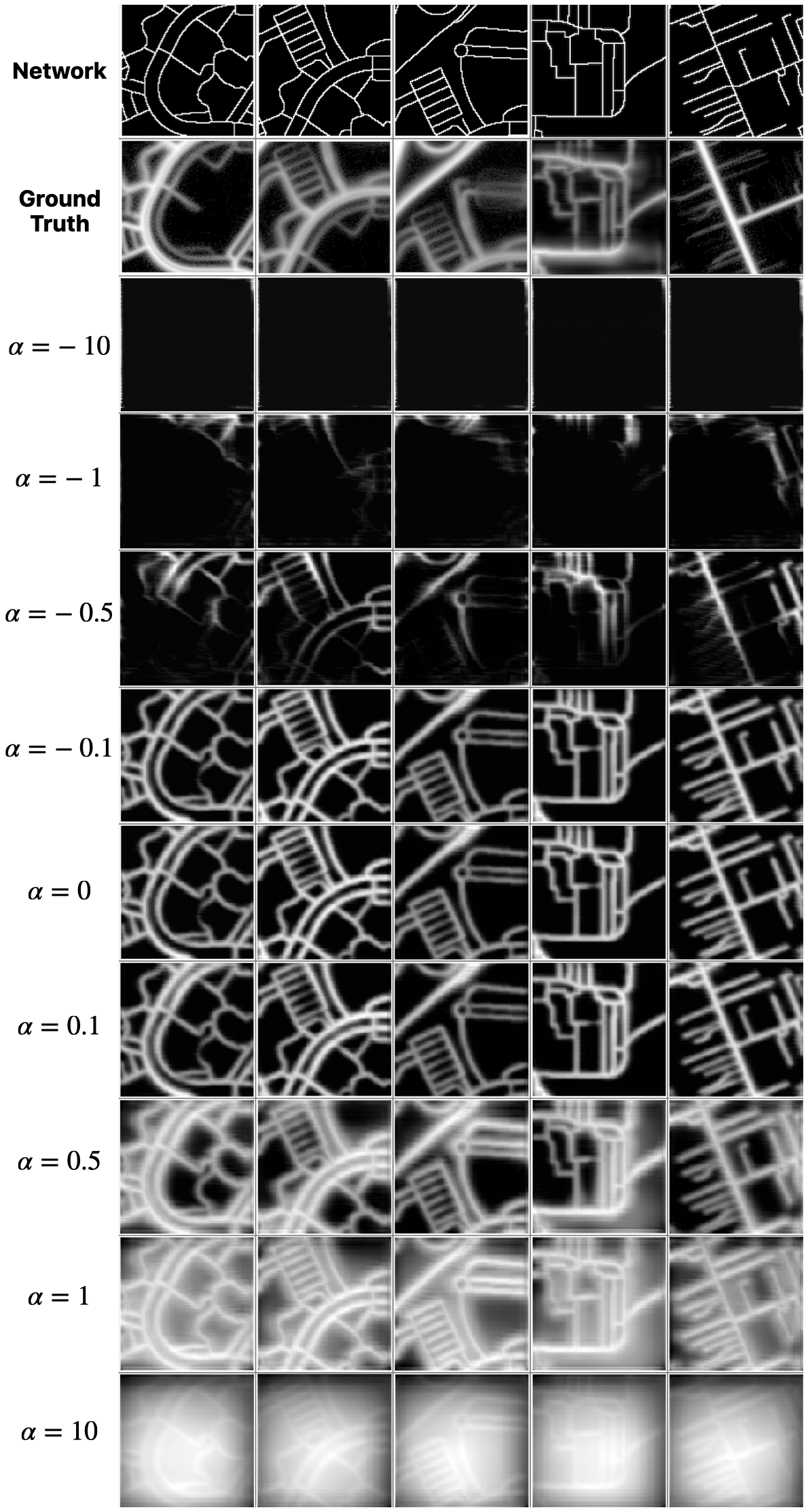}
        \caption{Tuning $\alpha$ for generating CRMs}
        \label{tuning_density}
    \end{subfigure}
    \hfill
    \begin{subfigure}{0.55\textwidth}
        \centering
        \includegraphics[width=\textwidth]{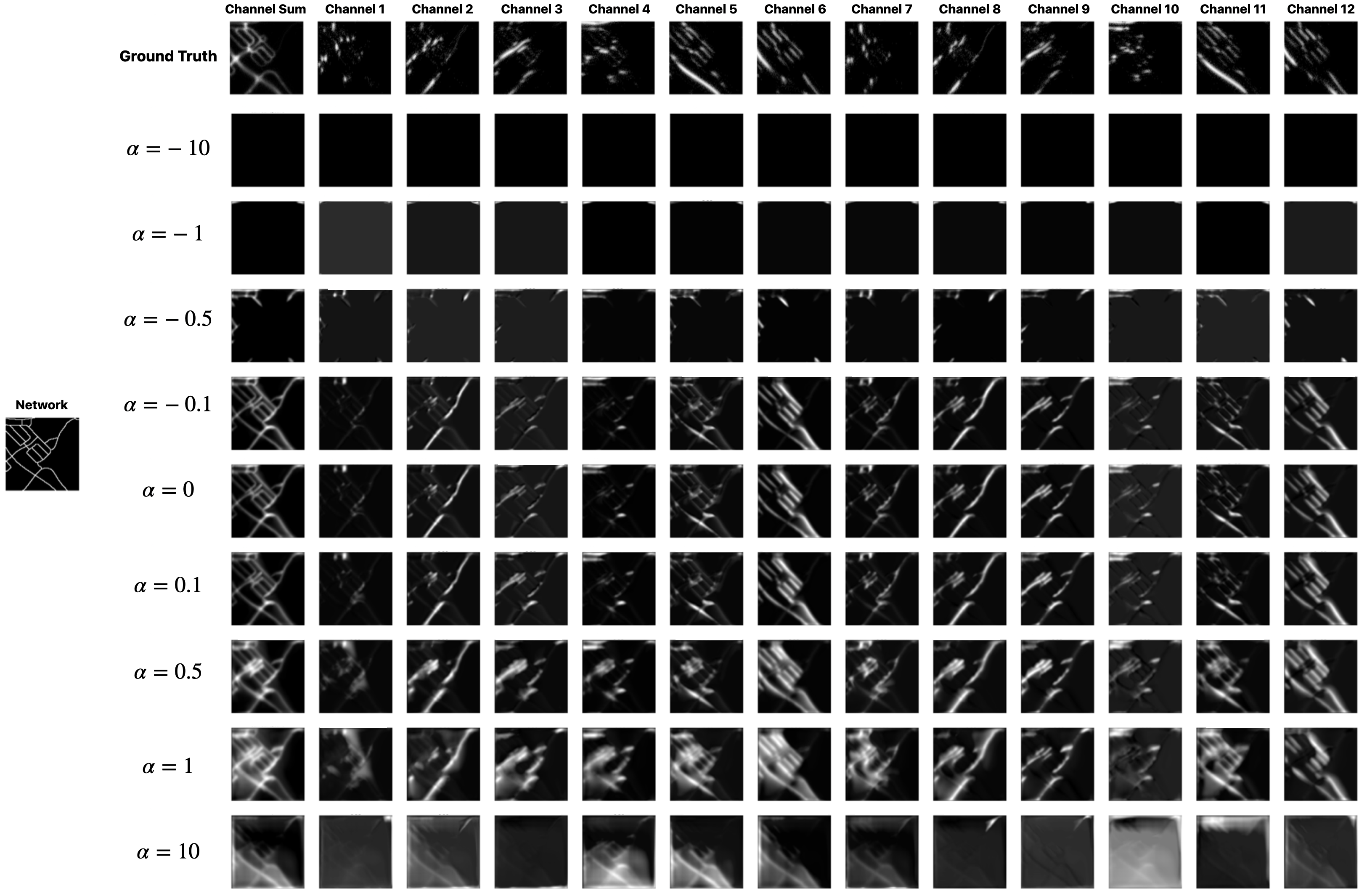}
        \caption{Tuning $\alpha$ for generating HCRM}
        \label{tuning_heading}
    \end{subfigure}
    \caption{(a) Five examples demonstrating the tuning of the latent representation by varying $\alpha$ for generating CRM. (b) An example demonstrating the tuning of the latent representation by varying $\alpha$ for generating HCRM.}
    \label{generated_tuning}
\end{figure*}

Figure \ref{tuning_density} shows the generation results for single binary attribute tuning for CRM, and figure \ref{tuning_heading} shows the same for HCRM, by varying $\alpha$ from -10 to 10. They suggest that our manipulation approach performs well on the attribute $\boldsymbol{a}$. In fact, the value of $\alpha$ corresponding to a desired probe density increase (or decrease) can be calibrated on the validation set. We found that when the DCGM was trained on data from the representative time interval, $\Delta t$, in the geographical region used to build the training, validation, and test sets, $\alpha = 0.45$ corresponds to the estimate for the probe aggregates if observations were made for data from $2 \Delta t$.

\section{Discussion of Results, Limitations, and Conclusions}\label{discussion_and_conclusion}
In this paper, we proposed a deep conditional generative model, called VAE-Info-cGAN, that simultaneously combines PLC and FLC to generate an image from a target distribution. The PLC is embedded into a vector space using the condition encoder to produce a condition vector. The feature-level condition is provided as an attribute vector. The proposed DCGM is shown to accurately generate image-like representations of spatio-temporal probe aggregates. Sections \ref{exp} and \ref{results} summarize the experiments conducted on the architecture of the DCGM and discuss the quality of the resulting generated images, respectively. Figures \ref{density_heatmaps} and \ref{heading_heatmaps} demonstrate that good quality (quantitative and qualitative) samples can be produced using this model. 

The design choice of exposing the latent space pays rich dividends as shown by the experiments on single attribute editing with the methodology described in section \ref{latent_variable_tuning}. Figures \ref{tuning_density} and \ref{tuning_heading} demonstrate this approach for the macro-level attribute of time duration of observation for aggregation of probe data. 

The proposed architecture of the model ensures that once the model has been trained and the latent space identified, diverse samples can be generated from the target distribution by simply modulating either one or both of the conditional inputs. Since these modulations can be done programmatically (e.g. automate adding or removing road segments from binary road network), efficient sampling from the target distribution is possible. Generating samples using a trained model is fast --- in our experiments, one forward pass of the inference computation graph (see figure \ref{architecture}(a)) for a batch of 32 examples takes (on average) 0.03s for CRM and 0.07s for HCRM on a NVIDIA Tesla V100 GPU. 

It must be emphasized that the PLC for all the generated images in this paper is only the binary road network containing segments which are not drivable too; which is sometimes evident in the cases where the ground truth contains no probes but the generation is guided by the road network. The design of VAE-Info-cGAN however, allows the user to guide the generation with more information. For example, if the model is trained with a two channel pixel-wise input where the first channel is the binary road network and the second channel is a binary mask signifying to the generator whether to populate the segments in the road network with probes or not (i.e., a driving restriction mask), the generator will learn to produce an output on drivable road segments only. Similarly, additional channels signifying direction of travel (e.g. one-way, two-way), form-of-way (e.g. pedestrian walkways, driveways, bike paths, etc.) may be provided to guide the generation of images in a more informative way. The proposed model design allows the user to make the generation as informative as required by the user's target application. 

Applications of the proposed generative model are wide and include automated synthetic data and labels generation for training supervised models, adversarial active learning for robust training of supervised models, data augmentation, and curated rare and extreme example generation, among others. In physics-based applications of deep generative models, having access to the latent representation can be exploited to encode physical constraints. For geospatial applications relevant to mapping demonstrated in this paper, this could mean encoding traffic rules and traffic flow constraints into the latent space for more accurate and informative generation of samples. This is a line of research that we intend to pursue in the future. 

Limitations that apply to similar generative models also hold for the proposed model. One of them is that the model inherits the characteristics of the training data. Thus, if we trained the model on data in Canada during summer and tried to predict the probe aggregates in winter when it snows heavily, there will be a large disparity between the ground truth and the predicted images since the model does not have knowledge of seasonal changes. However, the model may be explicitly trained to use seasonal information represented as a portion of the attribute vector $\boldsymbol{a}$ if the training data has examples from different seasonal distributions. The proposed latent representation tuning methodology can be used to learn the decision boundary that separates summer predictions from those in winter. One can then vary the vector $\boldsymbol{a}$ along the direction of the normal vector to the decision boundary and generate samples with seasonal variations. Similarly, the attribute vector, $\boldsymbol{a}$, can be used to encode a variety of macro-level features such as urban-rural differences in traffic, motion-modality differences in traffic (e.g., driving, walking), vehicular differences in traffic (e.g., cars, motorcycles, trucks), to name a few. The freedom given to the user of the proposed model thus leaves many unexplored possibilities that we aim to pursue in the future. 
\begin{figure*}[!htbp]
    \centering
    \includegraphics[width=0.65\textwidth]{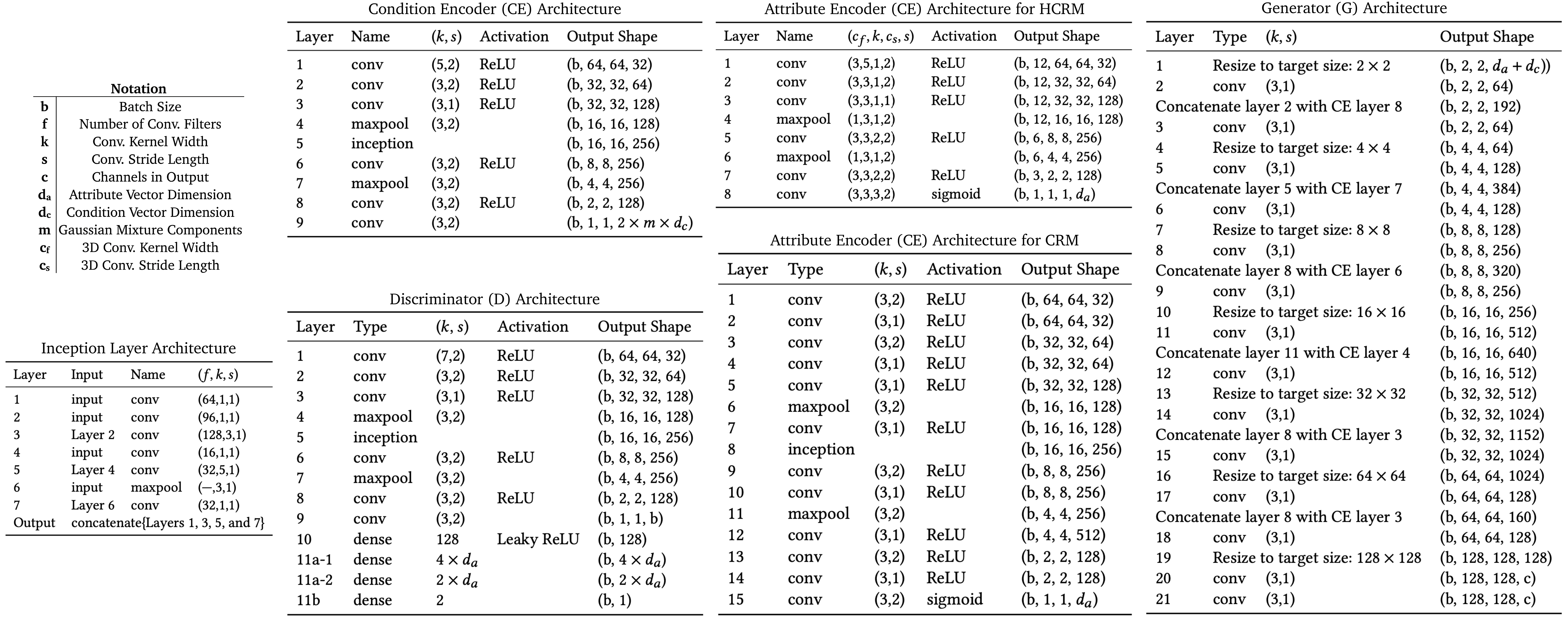}
    \caption{Details of neural architecture of components of VAE-Info-cGAN.}
    \label{neural_architecture}
\end{figure*}

\bibliographystyle{ACM-Reference-Format}
\bibliography{Paper_ACMSIGSPATIAL2020}


\begin{thebibliography}{29}


\ifx \showCODEN    \undefined \def \showCODEN     #1{\unskip}     \fi
\ifx \showDOI      \undefined \def \showDOI       #1{#1}\fi
\ifx \showISBNx    \undefined \def \showISBNx     #1{\unskip}     \fi
\ifx \showISBNxiii \undefined \def \showISBNxiii  #1{\unskip}     \fi
\ifx \showISSN     \undefined \def \showISSN      #1{\unskip}     \fi
\ifx \showLCCN     \undefined \def \showLCCN      #1{\unskip}     \fi
\ifx \shownote     \undefined \def \shownote      #1{#1}          \fi
\ifx \showarticletitle \undefined \def \showarticletitle #1{#1}   \fi
\ifx \showURL      \undefined \def \showURL       {\relax}        \fi
\providecommand\bibfield[2]{#2}
\providecommand\bibinfo[2]{#2}
\providecommand\natexlab[1]{#1}
\providecommand\showeprint[2][]{arXiv:#2}

\bibitem[\protect\citeauthoryear{Bowles et~al\mbox{.}}{Bowles
  et~al\mbox{.}}{2018}]%
        {bowles2018gan}
\bibfield{author}{\bibinfo{person}{C. Bowles} {et~al\mbox{.}}}
  \bibinfo{year}{2018}\natexlab{}.
\newblock \showarticletitle{{GAN augmentation: Augmenting training data using
  Generative Adversarial Networks}}.
\newblock \bibinfo{journal}{\emph{arXiv preprint arXiv:1810.10863}}
  (\bibinfo{year}{2018}).
\newblock


\bibitem[\protect\citeauthoryear{Brock et~al\mbox{.}}{Brock
  et~al\mbox{.}}{2018}]%
        {biggan}
\bibfield{author}{\bibinfo{person}{A. Brock} {et~al\mbox{.}}}
  \bibinfo{year}{2018}\natexlab{}.
\newblock \showarticletitle{{Large scale GAN training for high fidelity natural
  image synthesis}}.
\newblock \bibinfo{journal}{\emph{{arXiv preprint arXiv:1809.11096}}}
  (\bibinfo{year}{2018}).
\newblock


\bibitem[\protect\citeauthoryear{Burgess et~al\mbox{.}}{Burgess
  et~al\mbox{.}}{2018}]%
        {betaVAE}
\bibfield{author}{\bibinfo{person}{C.~P. Burgess} {et~al\mbox{.}}}
  \bibinfo{year}{2018}\natexlab{}.
\newblock \showarticletitle{{Understanding disentangling in $\beta$-VAE}}.
\newblock \bibinfo{journal}{\emph{arXiv preprint arXiv:1804.03599}}
  (\bibinfo{year}{2018}).
\newblock


\bibitem[\protect\citeauthoryear{Chan et~al\mbox{.}}{Chan
  et~al\mbox{.}}{2019}]%
        {chan2019everybody}
\bibfield{author}{\bibinfo{person}{C. Chan} {et~al\mbox{.}}}
  \bibinfo{year}{2019}\natexlab{}.
\newblock \showarticletitle{{Everybody dance now}}. In
  \bibinfo{booktitle}{\emph{Proceedings of the IEEE International Conference on
  Computer Vision}}. \bibinfo{pages}{5933--5942}.
\newblock


\bibitem[\protect\citeauthoryear{Chen et~al\mbox{.}}{Chen
  et~al\mbox{.}}{2016}]%
        {infogan}
\bibfield{author}{\bibinfo{person}{X. Chen} {et~al\mbox{.}}}
  \bibinfo{year}{2016}\natexlab{}.
\newblock \showarticletitle{{InfoGAN: Interpretable representation learning by
  information maximizing Generative Adversarial Nets}}.
\newblock


\bibitem[\protect\citeauthoryear{Engel et~al\mbox{.}}{Engel
  et~al\mbox{.}}{2019}]%
        {music}
\bibfield{author}{\bibinfo{person}{J. Engel} {et~al\mbox{.}}}
  \bibinfo{year}{2019}\natexlab{}.
\newblock \showarticletitle{{GANS}ynth: Adversarial neural audio synthesis}.
\newblock \bibinfo{journal}{\emph{arXiv preprint arXiv:1902.08710}}.
\newblock


\bibitem[\protect\citeauthoryear{Gal et~al\mbox{.}}{Gal et~al\mbox{.}}{2017}]%
        {gal2017deep}
\bibfield{author}{\bibinfo{person}{Y. Gal} {et~al\mbox{.}}}
  \bibinfo{year}{2017}\natexlab{}.
\newblock \showarticletitle{{Deep Bayesian active learning with image data}}.
\newblock \bibinfo{journal}{\emph{arXiv preprint arXiv:1703.02910}}.
\newblock


\bibitem[\protect\citeauthoryear{Ganguli, Garzon, and Glaser}{Ganguli
  et~al\mbox{.}}{2019}]%
        {ganguli2019geogan}
\bibfield{author}{\bibinfo{person}{S. Ganguli}, \bibinfo{person}{P. Garzon},
  {and} \bibinfo{person}{N. Glaser}.} \bibinfo{year}{2019}\natexlab{}.
\newblock \showarticletitle{{GeoGAN: A conditional GAN with reconstruction and
  style loss to generate standard layer of maps from satellite images}}.
\newblock \bibinfo{journal}{\emph{arXiv preprint arXiv:1902.05611}}
  (\bibinfo{year}{2019}).
\newblock


\bibitem[\protect\citeauthoryear{Kaneko et~al\mbox{.}}{Kaneko
  et~al\mbox{.}}{2017}]%
        {filtered}
\bibfield{author}{\bibinfo{person}{T. Kaneko} {et~al\mbox{.}}}
  \bibinfo{year}{2017}\natexlab{}.
\newblock \showarticletitle{{Generative attribute controller with conditional
  filtered Generative Adversarial Networks}}. In
  \bibinfo{booktitle}{\emph{Proceedings of the IEEE Conference on Computer
  Vision and Pattern Recognition}}. \bibinfo{pages}{6089--6098}.
\newblock


\bibitem[\protect\citeauthoryear{Karras et~al\mbox{.}}{Karras
  et~al\mbox{.}}{2018}]%
        {style}
\bibfield{author}{\bibinfo{person}{T. Karras} {et~al\mbox{.}}}
  \bibinfo{year}{2018}\natexlab{}.
\newblock \showarticletitle{{A style-based generator architecture for
  Generative Adversarial Networks}}.
\newblock \bibinfo{journal}{\emph{arXiv preprint arXiv:1812.04948}}.
\newblock


\bibitem[\protect\citeauthoryear{Kumar et~al\mbox{.}}{Kumar
  et~al\mbox{.}}{2017}]%
        {kumar2017variational}
\bibfield{author}{\bibinfo{person}{A. Kumar} {et~al\mbox{.}}}
  \bibinfo{year}{2017}\natexlab{}.
\newblock \showarticletitle{{Variational inference of disentangled latent
  concepts from unlabeled observations}}.
\newblock \bibinfo{journal}{\emph{arXiv preprint arXiv:1711.00848}}
  (\bibinfo{year}{2017}).
\newblock


\bibitem[\protect\citeauthoryear{Larsen et~al\mbox{.}}{Larsen
  et~al\mbox{.}}{2015}]%
        {vaegan}
\bibfield{author}{\bibinfo{person}{A.~B.~L. Larsen} {et~al\mbox{.}}}
  \bibinfo{year}{2015}\natexlab{}.
\newblock \showarticletitle{{Autoencoding beyond pixels using a learned
  similarity metric}}.
\newblock \bibinfo{journal}{\emph{arXiv preprint arXiv:1512.09300}}
  (\bibinfo{year}{2015}).
\newblock


\bibitem[\protect\citeauthoryear{Lee and Moloney}{Lee and Moloney}{2017}]%
        {8248284}
\bibfield{author}{\bibinfo{person}{K. Lee} {and} \bibinfo{person}{D. Moloney}.}
  \bibinfo{year}{2017}\natexlab{}.
\newblock \showarticletitle{{Evaluation of synthetic data for deep learning
  stereo depth algorithms on embedded platforms}}. In
  \bibinfo{booktitle}{\emph{4th International Conference on Systems and
  Informatics (ICSAI)}}. \bibinfo{pages}{170--176}.
\newblock


\bibitem[\protect\citeauthoryear{Ma et~al\mbox{.}}{Ma et~al\mbox{.}}{2019}]%
        {ma2019novel}
\bibfield{author}{\bibinfo{person}{Liqian Ma} {et~al\mbox{.}}}
  \bibinfo{year}{2019}\natexlab{}.
\newblock \showarticletitle{{A novel bilevel paradigm for image-to-image
  translation}}.
\newblock \bibinfo{journal}{\emph{arXiv preprint arXiv:1904.09028}}
  (\bibinfo{year}{2019}).
\newblock


\bibitem[\protect\citeauthoryear{Mirza and Osindero}{Mirza and
  Osindero}{2014}]%
        {cgans}
\bibfield{author}{\bibinfo{person}{M. Mirza} {and} \bibinfo{person}{S.
  Osindero}.} \bibinfo{year}{2014}\natexlab{}.
\newblock \showarticletitle{{Conditional Generative Adversarial Nets}}.
\newblock \bibinfo{journal}{\emph{arXiv preprint arXiv:1411.1784}}
  (\bibinfo{year}{2014}).
\newblock


\bibitem[\protect\citeauthoryear{Moreira-Matias et~al\mbox{.}}{Moreira-Matias
  et~al\mbox{.}}{2013}]%
        {ucigpsdataset}
\bibfield{author}{\bibinfo{person}{L. Moreira-Matias} {et~al\mbox{.}}}
  \bibinfo{year}{2013}\natexlab{}.
\newblock \showarticletitle{{Predicting taxi–passenger demand using streaming
  data}}.
\newblock \bibinfo{journal}{\emph{IEEE Transactions on Intelligent
  Transportation Systems}}  \bibinfo{volume}{14} (\bibinfo{year}{2013}),
  \bibinfo{pages}{1393--1402}.
\newblock


\bibitem[\protect\citeauthoryear{Nikolenko}{Nikolenko}{2019}]%
        {nikolenko2019synthetic}
\bibfield{author}{\bibinfo{person}{S.~I Nikolenko}.}
  \bibinfo{year}{2019}\natexlab{}.
\newblock \showarticletitle{{Synthetic data for deep learning}}.
\newblock \bibinfo{journal}{\emph{arXiv preprint arXiv:1909.11512}}
  (\bibinfo{year}{2019}).
\newblock


\bibitem[\protect\citeauthoryear{Odena et~al\mbox{.}}{Odena
  et~al\mbox{.}}{2016}]%
        {odena2016deconvolution}
\bibfield{author}{\bibinfo{person}{A. Odena} {et~al\mbox{.}}}
  \bibinfo{year}{2016}\natexlab{}.
\newblock \showarticletitle{{Deconvolution and checkerboard artifacts}}.
\newblock \bibinfo{journal}{\emph{Distill}} (\bibinfo{year}{2016}).
\newblock
\urldef\tempurl%
\url{http://distill.pub/2016/deconv-checkerboard}
\showURL{%
\tempurl}


\bibitem[\protect\citeauthoryear{Perez et~al\mbox{.}}{Perez
  et~al\mbox{.}}{2019}]%
        {perez2019semi}
\bibfield{author}{\bibinfo{person}{A. Perez} {et~al\mbox{.}}}
  \bibinfo{year}{2019}\natexlab{}.
\newblock \showarticletitle{{Semi-supervised multitask learning on
  multispectral satellite images using Wasserstein Generative Adversarial
  Networks (GANs) for predicting poverty}}.
\newblock \bibinfo{journal}{\emph{arXiv preprint arXiv:1902.11110}}
  (\bibinfo{year}{2019}).
\newblock


\bibitem[\protect\citeauthoryear{Registry}{Registry}{2020}]%
        {epsg_3857}
\bibfield{author}{\bibinfo{person}{EPSG Geodetic~Parameter Registry}.}
  \bibinfo{year}{2020}\natexlab{}.
\newblock \bibinfo{title}{{Official entry of EPSG:3857 spherical Mercator
  projection coordinate system (Date Accessed: Apr 16, 2020)}}.
\newblock \bibinfo{howpublished}{\url{http://www.epsg-registry.org}}.
\newblock


\bibitem[\protect\citeauthoryear{Research}{Research}{2011}]%
        {msrtaxi}
\bibfield{author}{\bibinfo{person}{Microsoft Research}.}
  \bibinfo{year}{2011}\natexlab{}.
\newblock \bibinfo{title}{{T-Drive trajectory data sample}}.
\newblock
\newblock
\urldef\tempurl%
\url{https://www.microsoft.com/en-us/research/publication/t-drive-trajectory-data-sample/}
\showURL{%
\tempurl}


\bibitem[\protect\citeauthoryear{Shen et~al\mbox{.}}{Shen
  et~al\mbox{.}}{2019}]%
        {interface}
\bibfield{author}{\bibinfo{person}{Y. Shen} {et~al\mbox{.}}}
  \bibinfo{year}{2019}\natexlab{}.
\newblock \showarticletitle{{Interpreting the latent space of GANs for semantic
  face editing}}.
\newblock \bibinfo{journal}{\emph{arXiv preprint arXiv:1907.10786}}
  (\bibinfo{year}{2019}).
\newblock


\bibitem[\protect\citeauthoryear{Sohn et~al\mbox{.}}{Sohn
  et~al\mbox{.}}{2015}]%
        {cvaes}
\bibfield{author}{\bibinfo{person}{K. Sohn} {et~al\mbox{.}}}
  \bibinfo{year}{2015}\natexlab{}.
\newblock \showarticletitle{{Learning structured output representation using
  deep conditional generative models}}.
\newblock In \bibinfo{booktitle}{\emph{Advances in Neural Information
  Processing Systems}}. \bibinfo{pages}{3483--3491}.
\newblock


\bibitem[\protect\citeauthoryear{Szegedy et~al\mbox{.}}{Szegedy
  et~al\mbox{.}}{2017}]%
        {szegedy2017inception}
\bibfield{author}{\bibinfo{person}{C. Szegedy} {et~al\mbox{.}}}
  \bibinfo{year}{2017}\natexlab{}.
\newblock \showarticletitle{{Inception-v4, Inception-Resnet, and the Impact of
  Residual Connections on Learning}}. In \bibinfo{booktitle}{\emph{31st AAAI
  conference on AI}}.
\newblock


\bibitem[\protect\citeauthoryear{TechCrunch}{TechCrunch}{2018}]%
        {tcrunch}
\bibfield{author}{\bibinfo{person}{TechCrunch}.}
  \bibinfo{year}{2018}\natexlab{}.
\newblock \bibinfo{title}{{Apple is rebuilding Maps from the ground up}}.
\newblock
\newblock
\urldef\tempurl%
\url{https://techcrunch.com/2018/06/29/apple-is-rebuilding-maps-from-the-ground-up/}
\showURL{%
\tempurl}


\bibitem[\protect\citeauthoryear{Theis et~al\mbox{.}}{Theis
  et~al\mbox{.}}{2015}]%
        {theis2015note}
\bibfield{author}{\bibinfo{person}{L. Theis} {et~al\mbox{.}}}
  \bibinfo{year}{2015}\natexlab{}.
\newblock \showarticletitle{{A note on the evaluation of generative models}}.
\newblock \bibinfo{journal}{\emph{arXiv preprint arXiv:1511.01844}}
  (\bibinfo{year}{2015}).
\newblock


\bibitem[\protect\citeauthoryear{Xie et~al\mbox{.}}{Xie et~al\mbox{.}}{2019}]%
        {xie2019unsupervised}
\bibfield{author}{\bibinfo{person}{Q. Xie} {et~al\mbox{.}}}
  \bibinfo{year}{2019}\natexlab{}.
\newblock \showarticletitle{{Unsupervised data augmentation}}.
\newblock \bibinfo{journal}{\emph{arXiv preprint arXiv:1904.12848}}
  (\bibinfo{year}{2019}).
\newblock


\bibitem[\protect\citeauthoryear{Zhang et~al\mbox{.}}{Zhang
  et~al\mbox{.}}{2017}]%
        {text}
\bibfield{author}{\bibinfo{person}{Y. Zhang} {et~al\mbox{.}}}
  \bibinfo{year}{2017}\natexlab{}.
\newblock \showarticletitle{{Adversarial feature matching for text
  generation}}. In \bibinfo{booktitle}{\emph{Proceedings of the 34th
  International Conference on Machine Learning}}. JMLR,
  \bibinfo{pages}{4006--4015}.
\newblock


\bibitem[\protect\citeauthoryear{Zhu et~al\mbox{.}}{Zhu et~al\mbox{.}}{2017}]%
        {cyclegan}
\bibfield{author}{\bibinfo{person}{J.~Y. Zhu} {et~al\mbox{.}}}
  \bibinfo{year}{2017}\natexlab{}.
\newblock \showarticletitle{{Unpaired image-to-image translation using
  cycle-consistent adversarial networks}}.
\newblock \bibinfo{journal}{\emph{arXiv preprint arXiv:1703.10593}}
  (\bibinfo{year}{2017}).
\newblock


\end{thebibliography}

\appendix

\section{Case Study: Detecting Temporal Changes in Traffic Patterns}\label{appendix_temporal_changes}
Given specific image inputs, $\boldsymbol{x}(t_1), \boldsymbol{x}(t_2) \in \mathbb{R}^{d \times d \times c}$, that correspond to encoding the traffic flow pattern using data from sensors or GPS trace trajectories at a geographic location at two different times ($t_1$ and $t_2$), consider training a model that is tasked with identifying locations where significant meaningful changes to the road network or traffic flow pattern have occurred in the given time interval $t_2-t_1$, e.g., road closure, junction changed to roundabout. Let the distribution of the input data be denoted as $s(\boldsymbol{x})$. A simple difference of the inputs, $\boldsymbol{x}(t_2) - \boldsymbol{x}(t_1)$, is not sufficient to detect these changes since the changes may result from noise, seasonal effects (winter versus summer traffic changes), etc. Cast as a semantic segmentation task, this application requires a class-balanced, labeled dataset that is challenging to create manually for the reasons mentioned in paragraph 3 of the introduction. 

Consider an ideal scenario where we have access to a trained DCGM that can synthetically generate $\boldsymbol{x}$ by sampling from $s(\boldsymbol{x})$ conditioned on an input road network along with some macro-level characteristics (e.g., season of the year, nature of terrain). Given a real example of input data, $\boldsymbol{x}^j$, we can utilize the trained DCGM to synthetically generate a large and diverse set of possible traffic patterns encoded as images, $\{\boldsymbol{\tilde{x}}_1^j, \boldsymbol{\tilde{x}}_2^j, \ldots, \boldsymbol{\tilde{x}}_k^j\}$, either by modifying the associated road network (e.g., adding or removing road segments, replacing junctions with roundabouts), or by varying both the macro-level conditions along with modifying the road network. The associated labels, $\{\boldsymbol{y}_1^j, \boldsymbol{y}_2^j, \ldots , \boldsymbol{y}_k^j\}$, are simply the locations where the road network is modified. This process can be repeated for all $N$ available examples of real data, $\{\boldsymbol{x}^1, \boldsymbol{x}^2, \ldots, \boldsymbol{x}^N\}$, which results in a synthetic dataset of size $Nk$, assuming each real example produces $k$ synthetically generated examples. 

A supervised image segmentation model (SISM) can then be trained to identify locations of changes such that the input to the model is the concatenation of $(\boldsymbol{x}^j,\boldsymbol{\tilde{x}}_i^j)$ and the label is $\boldsymbol{y}_i^j$ for $i = 1, 2, \ldots , k$ and $j = 1, 2, \ldots , N$. Since the dataset is augmented synthetically using the DCGM, any number of diverse examples can be generated on-demand allowing the user to pay specific attention to rare and extreme cases so that the training set has enough examples of the different possible scenarios that the SISM is being trained to identify. In other words, even if there is a bound on $N$ due to any of the reasons discussed in paragraph 3 of the introduction, $Nk$ can be arbitrarily large. 

Furthermore, qualitative and quantitative control on the generated image is of paramount importance and the quality of the generated samples influences the performance of the downstream models trained on the hybrid dataset. For example, given the characteristics of traffic flow at a junction of roads (at a particular time of the year and assuming the observations are made for a time interval $\Delta \tau$), one may want to synthetically generate the traffic flow pattern at the same location (for the same time of the year and for the same time interval $\Delta \tau$) for a variety of scenarios, e. g., one of the exits from the junction is closed, the junction is replaced by a roundabout, time interval is doubled to $2\Delta t$.

\section{Neural Architecture of Model}\label{appendix_neural_architecture}
Details of the neural architecture of the VAE-Info-cGAN's components used to generate results presented in this paper are shown in figure \ref{neural_architecture}. All components are designed to be fully convolutional. The inception layer follows \cite{szegedy2017inception}. $c=1$ for CRM, and $c=12$ for HCRM. Zero-padding is used to obtain the required output shapes in all convolution and pooling layers. All convolutional layers are followed by a batch normalization layer and ReLU activation function except when it is the last layer. The condition encoder accepts inputs with dimensions $(b,128,128,1)$ since the input, $\boldsymbol{y}$, is simply the binary road network which is a single channel image ($u = 1$). The attribute encoder architecture used for CRM uses 2D convolutions while the one for HCRM uses 3D convolutions. The last layer of the encoder of the AE component uses sigmoid activation to obtain $\boldsymbol{a}$ within the range $[0, 1]$. The convolutional layer 21 in the generator is directly followed by the softplus activation function to ensure positive outputs. In the discriminator, dense layer 10 is followed by a batch normalization layer. Layers 11a-1 and 11a-2 compose the auxiliary neural network $Q$. Both layers 11a-1 and 11b have layer 10 as the input. Layer 11b is the output logit corresponding to the discriminator's estimate that its input is real. The optimal values of $m$, $d_a$, and $d_c$ were found to be $m=20$ and $d_a = d_c = 32$.

\end{document}